\documentclass[11pt]{article}

\usepackage[preprint]{acl}

\usepackage{times}
\usepackage{latexsym}
\usepackage[T1]{fontenc}
\usepackage[utf8]{inputenc}
\usepackage{microtype}
\usepackage{inconsolata}

\usepackage{hyperref}
\usepackage{url}
\usepackage[ruled,vlined]{algorithm2e}
\usepackage{graphicx}
\usepackage{xcolor}
\usepackage{amsfonts}
\usepackage{amsmath}
\usepackage{multirow}

\usepackage{subcaption}

\usepackage{stfloats}
\usepackage{caption}
\usepackage{cuted}
\usepackage{capt-of} 

\newcommand{\methodname}{CARES}
\usepackage{booktabs}
\usepackage{siunitx} 
\usepackage{xcolor, colortbl}
\usepackage{enumitem}
\definecolor{Gray}{gray}{0.9}
\definecolor{LightCyan}{rgb}{0.88,1,1}
\definecolor{golden}{rgb}{1,0.84, 0}
\title{CARES: Context-Aware Resolution Selector for VLMs}

\author{
  Moshe Kimhi$^{1,2}$\thanks{\ \ Equal contribution} \quad
  Nimrod Shabtay$^{2,3}$\footnotemark[1] \quad
  Raja Giryes$^{3}$ \quad
  Chaim Baskin$^{4}$\thanks{\ \ Equal supervision} \quad
  Eli Schwartz$^{2}$\footnotemark[2] \\
  $^{1}$Technion 
  $^{2}$IBM Research \\
  $^{3}$Tel-Aviv University \quad
  $^{4}$Ben-Gurion University \\
  \newline \textbf{Project Page:} \url{https://mkimhi.github.io/CARES/} \\
}

\begin{document}
\maketitle


\begin{abstract}
Large vision–language models (VLMs) commonly process images at native or high resolution to remain effective across tasks. This inflates visual tokens up to to 99\% of total tokens of the prefill stage, resulting in high compute and latency, even when low-resolution images would suffice. We introduce \emph{CARES}—a \textbf{C}ontext-\textbf{A}ware \textbf{R}esolution \textbf{S}elector, a lightweight preprocessing module that, given an image–query pair, predicts the \emph{minimal} sufficient input resolution. CARES uses a compact VLM (350M) to extract features and predict when a target pretrained VLM's response converges to its peak ability to answer correctly. Though trained as a discrete classifier over a set of optional resolutions, CARES interpolates continuous resolutions at inference for fine-grained control.
Across nine multimodal benchmarks spanning documents and natural images, as well as diverse target VLMs, CARES preserves task performance while reducing compute by up to $78\%$ on average across 9 benchmarks. 

\end{abstract}

\section{Introduction}
\begin{figure}
\centering
\includegraphics[width=\linewidth]{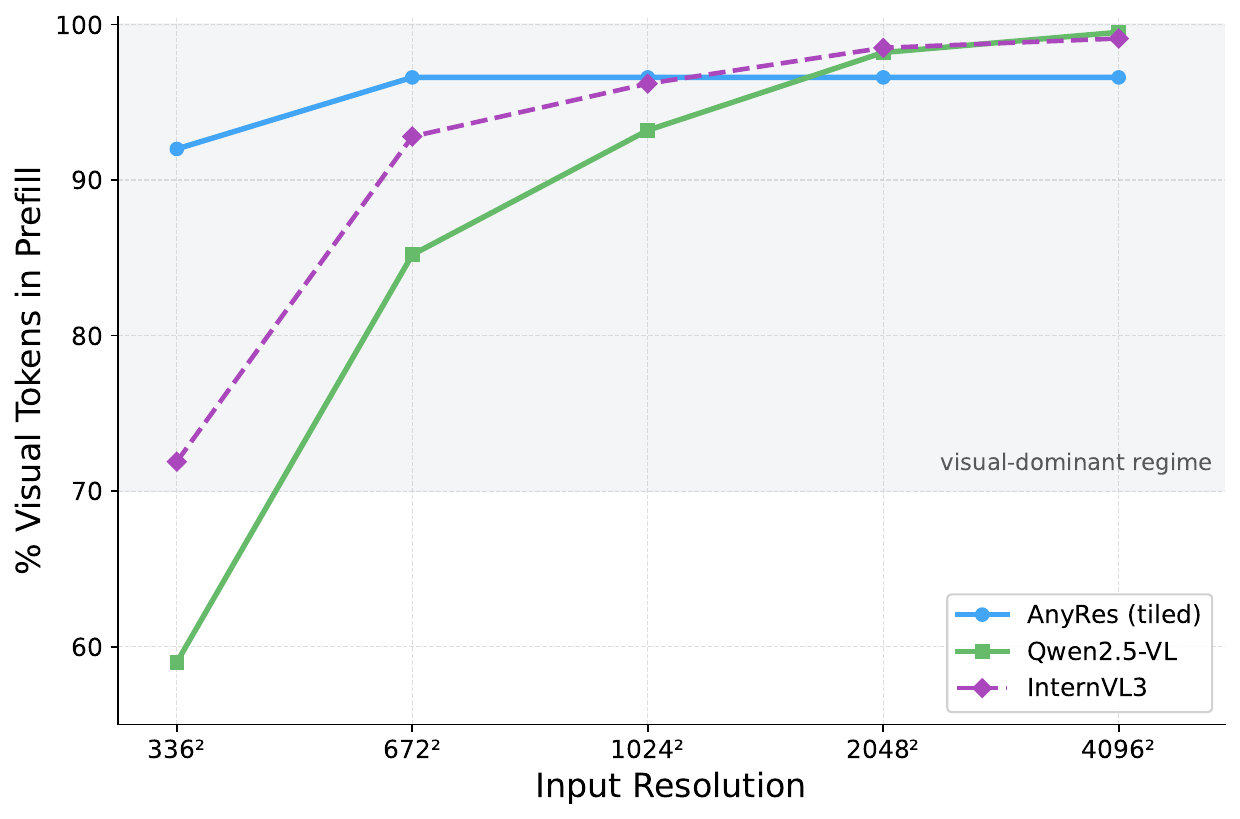}
\caption{\textbf{Visual token dominance across resolutions.} Fraction of visual tokens relative to a fixed 100-token text prompt. As resolution increases, visual tokens quickly dominate the context window, particularly in dynamic-resolution models where scaling is quadratic. AnyRes refer to tiling of multiple views. More details in Appx~\ref{app:token_count}.}
\label{fig:tokens}
\vspace{-5pt}
\end{figure}

Large vision–language models (VLMs) are increasingly used as general-purpose systems that solve a broad variety of visual tasks using a single model. Since the complexity and nature of each task are not known in advance, these models typically process images at very high resolutions to preserve the visual detail necessary for any potential query. This leads to a sharp increase in the number of visual tokens, as modern architectures map higher resolutions to proportionally more tokens. Strategies like AnyRes and tiling further increase token counts in order to capture fine-grained information \citep{liu2024llavanext,Qwen2VL}. In practical settings, visual tokens make up to 99\% of all tokens processed per request, which significantly impacts latency and memory consumption (Fig~\ref{fig:tokens}), even when the actual query may only require a coarse understanding of the scene.



\begin{figure*}
\centering
\includegraphics[width=\linewidth]{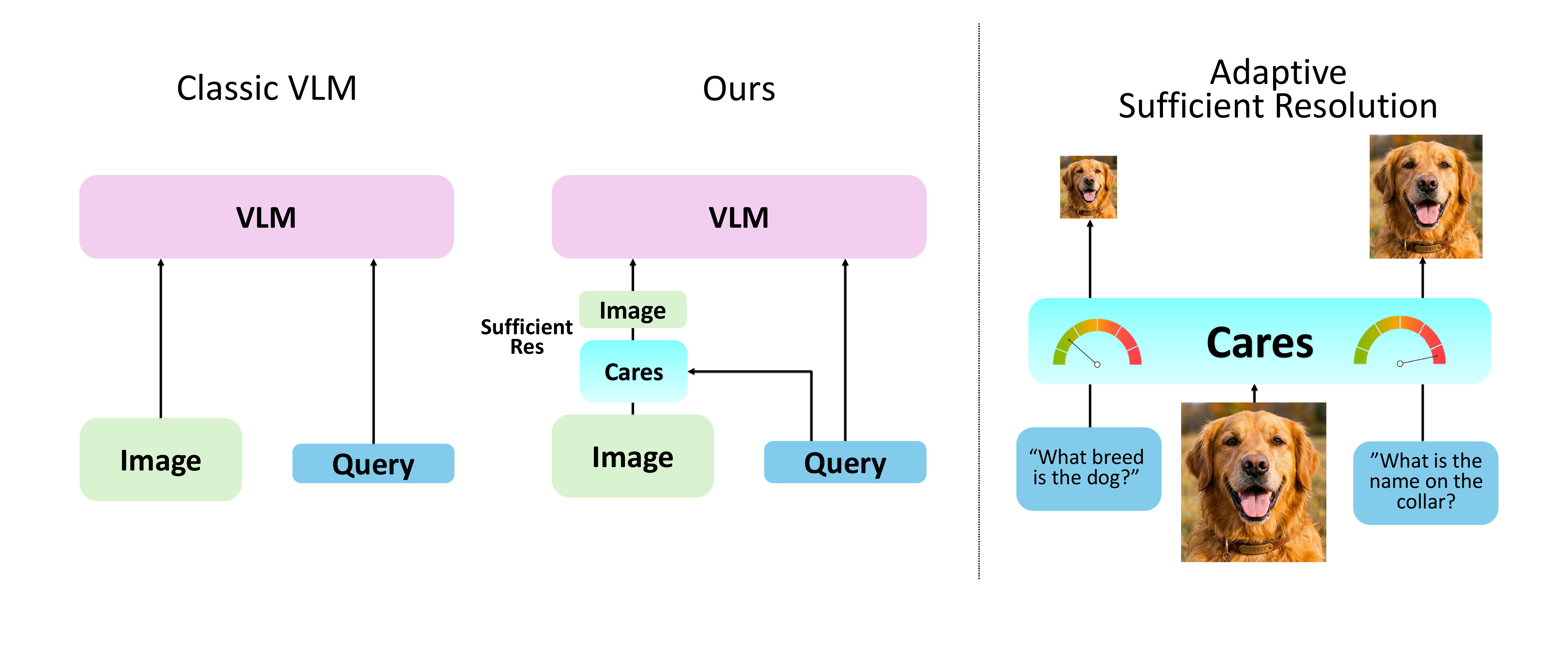}
\vspace{-30pt}
\caption{
Overview of \textbf{CARES}. On the left, we compare the traditional pipeline of a use of VLM vs the pipeline using \methodname{}. Given an image and its query, CARES predicts the minimal sufficient input resolution. The image is resized accordingly and, together with the query, passed to a downstream VLM. Coarse queries are routed to lower resolution; fine-grained queries that require more detail trigger higher resolution, which yields more visual tokens in the VLM.}
\label{fig:teaser}
\end{figure*}

A key observation is that \emph{not all queries require the same visual granularity}. Coarse queries (e.g., ``What is the breed of the dog?'') are typically answerable from a small image; fine‑grained queries (e.g., ``What is the name on the collar?'') benefit from higher resolution. 
Existing efficiency methods typically operate \emph{after} tokenization, on the output of the vision encoder -pruning, pooling, merging, or compressing with Q-former style architecture \citep{hired,sparsevlm,pyramiddrop,vtw,rao2021dynamicvit,evit,tome,tokenflex,cai2024matryoshka}.
While complementary, these methods typically operate on the output of the visual encoder alone and are unaware of the text input or the current query. Yet a more fundamental lever remains untouched: 
\emph{Can we choose the input granularity as a pre-processing step?}

We propose a \emph{Context‑Aware Resolution Selector} (\textbf{\methodname{}}), a lightweight model that, for a given image-query pair, selects the \emph{minimal} sufficient resolution to answer the query (Fig.~\ref{fig:teaser}). \methodname{} is model‑agnostic, placed \emph{in front of} an arbitrary VLM.
While our main instantiation uses a compact frozen VLM with a lightweight discriminative classifier, the CARES formulation is not tied to a specific predictor architecture. We also study a closely related autoregressive instantiation based on Granite-Docling, fine-tuned with LoRA, and report it separately on document-centric benchmarks.

It operates in three steps: 
\begin{itemize}
    \item A cheap low-resolution pass (e.g., $\leq 384^2$) extracts a joint image--query representation using a small proxy VLM.
    \item Given this representation, a lightweight classifier predicts the minimal resolution required for the task.
    \item The image is resized to the predicted resolution and passed to the target VLM. No changes to the VLM’s architecture, weights, or training are required.
\end{itemize}

A central challenge is supervision: what resolution is \emph{truly} sufficient for each example? 
We introduce a simple labeling procedure based on a discrete set of resolutions $\mathcal{R}$ and a task performance metric.
For each image, query, and GT response, we evaluate a pretrained VLM with increasingly higher resolution up to convergence in terms of the task metric (or reaching the native resolution).
The lowest resolution at which the convergence occurs is selected as the ground-truth optimal resolution for training \methodname{}.
Using a discrete resolution set avoids the cost of exhaustively searching over continuous values. Since the labels are discrete, the model is trained as a classifier. At inference time, however, we interpolate between the predicted class probabilities to recover a continuous resolution estimate.

Across 9 multimodal benchmarks, varying from natural images to document understanding (Section~\ref{sec:exp}) and different open and api-based model, \methodname{} reduces average visual tokens and GFLOPS by 70-80\%, with minimal to no accuracy drop compared to always using the highest (native) resolution.

\paragraph{Our contributions are as follows:}
\begin{enumerate}
    \item We define the task of \textit{query- and image-conditioned resolution selection} for vision-language models, aimed at reducing input size without sacrificing accuracy.
    
    \item We propose a simple yet effective supervision strategy based on multi-resolution rollouts and a convergence rule, yielding per-example sufficient resolution ground-truth, enabling training and evaluation.
    
    \item We introduce \methodname{}, a lightweight, model-agnostic module that selects resolution as a pre-processing step, requiring no changes to the target VLM.
    
    \item We demonstrate that many visual tokens are unnecessary: \methodname{} preserves performance across tasks while reducing visual compute by up to 78\% on average across 9 benchmarks, and is orthogonal with post-tokenization token compression.
\end{enumerate}

\section{Related Work}
\label{sec:related}

\paragraph{Visual-token sparsification at inference}
A growing line of work trims visual tokens \emph{after} tokenization inside the VLM stack. HiRED uses \texttt{[CLS]} attention to allocate a per-partition token budget and drop the least-informative vision tokens under a fixed budget, yielding large speedups on high-resolution inputs without retraining \citep{hired}. SparseVLM proposes a training-free, text-guided strategy: self-attention matrices rank visual tokens with an adaptive layer-wise sparsification ratio and a token-recycling mechanism to preserve information \citep{sparsevlm}. PyramidDrop stages the model and progressively reduces tokens at stage boundaries, motivated by the observation that redundancy increases with depth; it accelerates both training and inference and can also be used in a plug-and-play inference mode \citep{pyramiddrop}. Complementary to these, Visual Tokens Withdrawal (VTW) argues that visual information migrates to text tokens in early layers and thus withdraws vision tokens beyond a learned layer, cutting compute while maintaining quality \citep{vtw}. In contrast, \methodname{} decides \emph{before} tokenization which input resolution to use and leaves all VLM's components frozen.

\paragraph{Training for flexible token budgets}
TokenFLEX trains VLMs to operate across a range of visual–token counts by stochastically modulating tokens during training and adding a lightweight projector with adaptive pooling \citep{tokenflex}. \emph{Matryoshka Multimodal Models} (MMM) further pursue elastic compute, training nested representations that remain useful under progressively smaller token/feature budgets \citep{cai2024matryoshka}. \emph{LLaVA-Mini} pushes efficiency to the extreme by compressing visual information into (nearly) a single vision token while retaining competitive performance for both images and videos \citep{llava_mini}. \methodname{} targets the complementary axis of \emph{adaptive pixel allocation} before tokenization: it selects the minimal input resolution needed for a target utility and can front-end TokenFLEX/Matryoshka/LLaVA-Mini–style models to reduce pixels (and thus tokens) further.

\paragraph{Any-resolution inputs and tiling}
Many modern ViTs \cite{dehghani2023patch,beyer2023flexivitmodelpatchsizes} and VLMs boost fine-grained perception with AnyRes/dynamic-high-resolution tiling (e.g., LLaVA‑NeXT) or native dynamic resolution that maps larger images to more tokens (e.g., Qwen2‑VL) \citep{liu2024llavanext,Qwen2VL}. While effective, these strategies often increase visual tokens substantially. \methodname{} explicitly \emph{avoids} unnecessary tiling by routing easy cases to low resolutions and only escalating when the query and low-res cues predict a benefit.

\paragraph{Dynamic computation}
Vision-only methods reduce computation via token pruning/merging inside ViTs-e.g., DynamicViT prunes tokens hierarchically with learned importance \citep{rao2021dynamicvit}, EViT reorganizes/discards inattentive tokens \cite{evit}, and ToMe merges similar tokens on the fly \citep{tome}.WAVECLIP replaces patch tokenization with a multi-level wavelet tokenizer and performs coarse-to-fine inference in a single ViT \cite{kimhi2025wave}. For VLMs, SGL routes easy cases via a small ‘stitch’ model and defers hard ones to a larger counterpart, akin to early-exit routing \cite{zhao2024stitch}. These operate \emph{within} the encoder after tokenization; \methodname{} is complementary, deciding how many pixels to tokenize in the first place.

\paragraph{Adaptive input resolution selection}
Outside VLMs, dynamic-resolution networks learn a per-image resolution predictor that trades accuracy for cost in classification \citep{drnet}. CARES brings this idea to multimodal QA, conditions the policy on the query text, and supervises it with \emph{per-example} multi-resolution rollouts of the target VLM using a sufficiency rule, which yields unambiguous labels at deployment resolutions.

\paragraph{Extreme compression and design insights}
Recent analyses argue that, under fixed inference budgets, compute-optimal VLMs may prefer very few visual tokens and a larger LLM \citep{one_token}. Such results support approaches that minimize visual tokens when possible; methods like \emph{LLaVA-Mini} instantiate the “one-token vision’’ regime in practice \citep{llava_mini}. \methodname{} provides a query-conditioned mechanism to reduce pixels upstream, complementing these token-minimal designs.


\section{\methodname{}}
\label{sec:method}

This section outlines the problem addressed by \methodname{} (\ref{subsec::problem_setup}), followed by a description of the dataset generation procedure (\ref{subsec::dataset_generation}). We then detail the architecture and the training details of \methodname{} (\ref{subsec::arch}), Finally
 we outline our continuous resolution approach (\ref{subsec::continuous}).

\subsection{Problem Definition}
\label{subsec::problem_setup}
Given an image $x$ and query $q$, let $\mathcal{R} = [r_{\min}, r_{\max}] \subset \mathbb{R}^{+}$ denote the range of valid input resolutions and let $F$ be a fixed VLM. For any resolution $r \in \mathcal{R}$, we denote by $x^{(r)}$ the image $x$ resized such that its largest dimension equals $r$. Feeding $x^{(r)}$ and $q$ into $F$ yields an output $y = F(x^{(r)}, q)$. The VLM forms $T(r)$ visual tokens at resolution $r$ (including AnyRes/tiling effects). Our goal is to learn a \emph{selector} $f_\theta$ that predicts, from a single inexpensive low‑resolution pass at $r_{min}$, the minimal \emph{sufficient} resolution $r_s\in\mathcal{R}$ for accurately answering the query $q$ given image $x$.

\subsection{Labeling Strategy for Training CARES}
\label{subsec::dataset_generation}

Since searching for the optimal $r^{\star} \in \mathcal{R}$ is prohibitively expensive, we chose to use a small, discrete set of valid resolutions for the annotation $\mathcal{R}_d = \{r_1,...,r_K\} \subset \mathcal{R}$.
For each sample, we render the image at the fixed resolutions, $\mathcal{R}_d$, and use a pretrained VLM to generate predictions at each resolution. The predictions are evaluated against the ground-truth annotations using the ANLS metric. The supervision label is assigned as the lowest resolution whose ANLS score exceeds a threshold, without significant improvement at higher resolutions. The procedure yields a \emph{discrete} sufficiency label $r^{\star} \in \mathcal{R}_d$ per example. We emphasize that discretization is only used for supervision efficiency; at inference, we deploy a \emph{continuous} finer-grained selector (§\ref{subsec::continuous}). Algorithm \ref{alg:label} outlines the data generation process, and Table.~\ref{tab:data_pipeline} visualizes the concept.

\begin{table*}[ht]
\centering
\footnotesize
\begin{tabular}{m{1.5cm}m{4.3cm}m{4.3cm}m{4.3cm}}
&
\includegraphics[width=4.4cm]{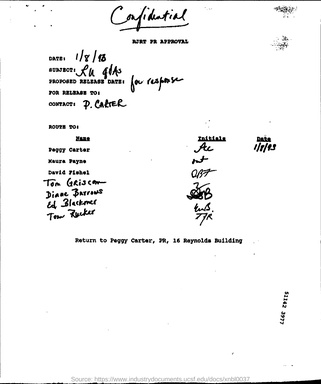} &
\includegraphics[width=4.4cm]{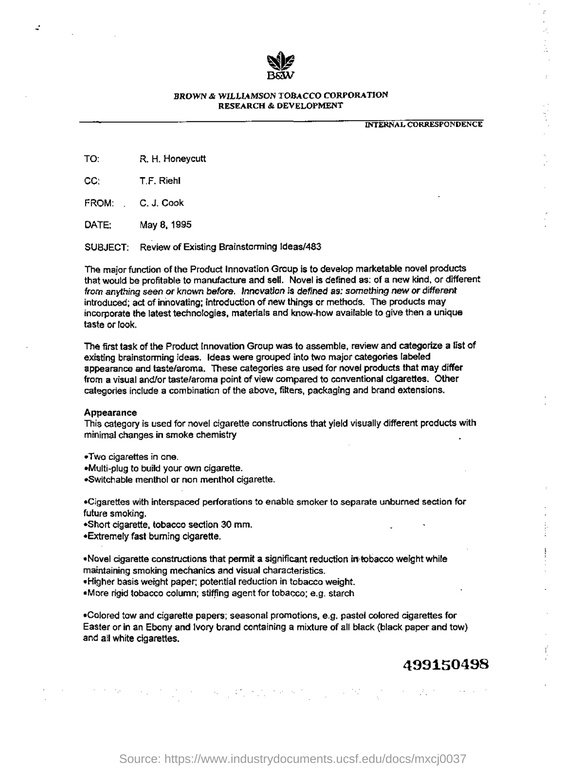} &
\includegraphics[width=4.4cm]{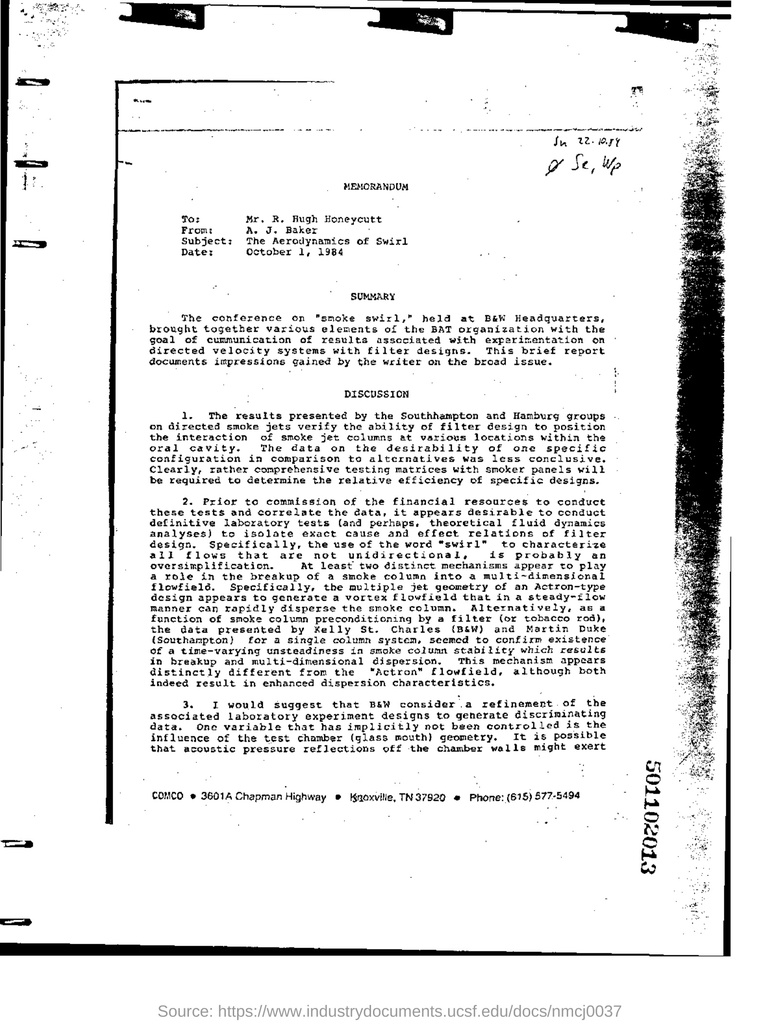} \\
\textbf{Query} &
what is the contact person name mentioned in letter? &
Who is in cc in this letter? &
One variable that has implicitly not been controlled? \\
\hline
\textbf{GT} &
P. Carter &
T.F. Riehl &
influence of the test chamber (glass mouth) geometry. \\
\hline
\textbf{Resp@384} &
P. Carter  &
T.F. Rosel  &
concentration of the final product \\
\textbf{ANLS} &
1.0 &
0.7 &
0.0 \\
\hline
\textbf{Resp@768} &
 &
T.F. Riehl  &
the influence of the test chamber (i.e. ash seath) geometry on the flow  \\
\textbf{ANLS} &
 &
1.0 &
0.65 \\
\hline
\textbf{Resp@1024} &
 &
 &
the influence of the test chamber (glass mouth) geometry. \\
\textbf{ANLS} &
 &
 &
0.93 \\
\hline
\textbf{Sufficient Resolution} &
384 &
768 &
1024 \\
\hline
\end{tabular}
\caption{Data generation pipeline for training \methodname{}. We process each input through a pretrained VLM (Granite-Vision) at three fixed resolutions and select the smallest resolution that produces a sufficient answer quality according to the ANLS metric.}
\label{tab:data_pipeline}
\end{table*}

Formally, we compute the ANLS score for each resolution:
\begin{equation}
u_k=\text{ANLS}\!\left(F(x^{(r_k)},q),\text{gt}\right)\!\in[0,1]
\label{eq:anls}
\end{equation}
and select the minimal sufficient resolution as:
\begin{equation}
r^\star=\min\Big\{r_k\;\Big|\; u_k\ge\tau,\ \max_{\ell> k}(u_\ell-u_k)\le \delta \Big\}
\label{eq:sufficiency}
\end{equation}
where we default to $r_K$ if no resolution satisfies the condition. 
We set $\tau{=}\,0.85$ and use a small margin $\delta$ (e.g., $0.1$) to prevent rewarding negligible performance improvements.
We define the full resolution range as $\mathcal{R} = [384, 1024]$, and use a discrete set $\mathcal{R}_d = \{384,768,1024\}$ for annotation.

\begin{algorithm}[t]
\DontPrintSemicolon
\SetKwInput{KwInput}{Input}
\SetKwInput{KwOutput}{Output}
\KwInput{$(x,q)$; resolutions $\mathcal{R}$; VLM $F$; utility $U$; threshold $\tau$; margin $\delta$}
\KwOutput{Label $r^\star\!\in\!\mathcal{R}$}
\For{$k\leftarrow 1$ \KwTo $K$}{
  $y_k\leftarrow F(x^{(r_k)},q)$;\ $u_k\leftarrow U(y_k,\text{gt})$\;
}
\For{$k\leftarrow 1$ \KwTo $K$}{
  \If{$u_k\ge\tau$ \textbf{and} $\max_{\ell> k}(u_\ell{-}u_k)\le\delta$}{
    \Return $r^\star\leftarrow r_k$
  }
}
\Return $r^\star\leftarrow r_K$
\caption{Labeling via multi‑resolution sufficiency rollouts.}
\label{alg:label}
\end{algorithm}

\subsection{Model Instantiations}
\label{subsec::arch}
Unless otherwise stated, all main experiments in this paper use the following discriminative instantiation of \methodname{}.

We design \methodname{} as a lightweight resolution selector that can be deployed in front of any vision–language model (VLM) to improve efficiency. Its behavior is governed by three core principles:

\begin{enumerate}
    \item \textbf{Compactness}: minimal overhead in computation and memory.
    \item \textbf{Preprocessing role}: determines resolution directly from raw inputs before invoking the VLM.
    \item \textbf{VLM-agnosticism}: works with any VLM, whether run locally or accessed via API, with no architecture changes or retraining required.
\end{enumerate}

\medskip
To implement these principles, we use a compact frozen VLM backbone as a joint vision–text feature extractor, followed by a lightweight classifier head.

Specifically, we adopt the pretrained SmolVLM-500M model \cite{marafioti2025smolvlm}, with layers 17--32 removed, as the backbone. Given an image at resolution $r_{\min}$ and a text query, we feed both into the model and extract the hidden state of the final token at layer 16. This representation encodes the joint image–query context and is passed to a classifier that outputs a soft distribution over target resolutions.
This design is motivated by recent findings showing that intermediate layer activations in LLMs and VLMs encode rich perceptual and semantic information that may not be surfaced at the output layer \citep{orgad2024llms,zhang2025mllms}. 
In addition to being more informative, as also evidenced by the performance gap in Table~\ref{tab:feature_extractor} where using intermediate features outperforms last-layer features by about 1\%, this choice substantially reduces computation since only roughly half of the LLM is used for feature extraction.

The resulting \methodname{} module has approximately 350M parameters and is trained with supervision over discrete resolution labels (see §\ref{subsec::dataset_generation}).

\paragraph{Autoregressive document-specialized instantiation.}
In addition to the discriminative selector above, we also instantiate \methodname{} using an autoregressive vision-language model. Concretely, we start from Granite-Docling-258M \cite{docling} and fine-tune it with LoRA (rank 8) on the same resolution-selection training set. Given the low-resolution image and the query, the model is prompted to predict one resolution label from the discrete set $\mathcal{R}_d=\{384,768,1024\}$. To avoid tokenization ambiguity, we map these labels to dedicated tokens \texttt{<1>}, \texttt{<2>} and \texttt{<3>}. 

At inference time, we read the first-step logits over the resolution tokens, apply a softmax to obtain class probabilities, and use the same expectation-based interpolation described in Eq.~\ref{eq:linear_expectation} to produce a continuous resolution. This preserves the deployment rule of CARES while replacing the discriminative head with an autoregressive predictor.

\subsection{From Discrete Supervision to a Continuous Resolution}\label{subsec::continuous}

Although \methodname{} is trained as a $K$-way classifier over a discrete set of resolutions $\mathcal{R}_d=\{r_1<\cdots<r_K\}$, we deploy it as a \emph{continuous} selector over $\mathcal{R}=[r_{min},r_{max}]$. 
Given features $z$ from the low-resolution image and query, compute logits $\ell(z) \in \mathbb{R}^K$ and class probabilities
\[
p=\mathrm{softmax}(\ell),
\]

We use the probability-weighted expectation over $\mathcal{R}_d$:
\begin{equation}
\tilde r \;=\; \sum_{k=1}^{|\mathcal{R}_d|} p_k\, r_k,
\label{eq:linear_expectation}
\end{equation}
This yields a \emph{continuous} resolution that varies smoothly with confidence and is insensitive to the specific discretization used for labeling. In practice, $\tilde r$ preserves the routing behavior of the classifier while allowing finer control.

\paragraph{Continuous inference algorithm.}
\vspace{-4pt}
\begin{algorithm}
\DontPrintSemicolon
\SetKwInput{KwInput}{Input}
\SetKwInput{KwOutput}{Output}
\KwInput{$(x,q)$; low-res $r_1$; logits $\ell$.}
\KwOutput{Continuous resolution $\tilde r \in [r_1,r_K]$.}
$z \leftarrow$ features from proxy VLM at $r_1$\;
$p \leftarrow \mathrm{softmax}( \ell(z))$\;
$\tilde r \leftarrow \sum_{k=1}^K p_k\, r_k$\; 
\Return $\tilde r$
\caption{\textbf{Continuous resolution selection}.}
\label{alg:continuous}
\end{algorithm}
\vspace{-8pt}

\paragraph{Deployment.}
The target VLM receives $x$ with the largest dimension resized to $\tilde r$ (or to the nearest supported side length to avoid under-allocation). For backbones that only accept a discrete set of input sizes, we round \emph{up} to the next supported size.
\begin{table*}[ht]
  \centering

  \resizebox{\textwidth}{!}{
  \begin{tabular}{
    l
    cc cc cc cc cc
    cc cc cc cc
    |cc
  }
    \toprule
      &
    \multicolumn{2}{c}{\textbf{Ai2D}} &
    \multicolumn{2}{c}{\textbf{ChartQA}} &
    \multicolumn{2}{c}{\textbf{DocVQA}} &
    \multicolumn{2}{c}{\textbf{OCRBench}} &
    \multicolumn{2}{c}{\textbf{SeedBench-2}} &
    \multicolumn{2}{c}{\textbf{MMMU}} &
    \multicolumn{2}{c}{\textbf{RealWorldQA}} &
    \multicolumn{2}{c}{\textbf{InfoVQA}} &
    \multicolumn{2}{c}{\textbf{MathVista}} &
    \multicolumn{2}{c}{\textbf{Average}}\\

    \cmidrule(lr){2-3}\cmidrule(lr){4-5}\cmidrule(lr){6-7}\cmidrule(lr){8-9}
    \cmidrule(lr){10-11}\cmidrule(lr){12-13}\cmidrule(lr){14-15}
    \cmidrule(lr){16-17}\cmidrule(lr){18-19}\cmidrule(lr){20-21}

    \textbf{Model} &
    {Score} & {Cost} &
    {Score} & {Cost} &
    {Score} & {Cost} &
    {Score} & {Cost} &
    {Score} & {Cost} &
    {Score} & {Cost} &
    {Score} & {Cost} &
    {Score} & {Cost} &
    {Score} & {Cost} &
    {Score} & {Cost} \\
    \midrule

    Granite-Vision -2B
      & 0.74 &  & 0.86 &  & 0.90 &  & 0.80 &  & 0.72 &
      & 0.29 &  & 0.17 &  & 0.35 &  & 0.48 &
      & 0.59 &  \\

    \rowcolor{golden!20}+ \textsc{\methodname{}}
      & 0.73 & -67\% & 0.87 & -69\% & 0.90 & -68\% & 0.80 & -68\% & 0.72 & -44\%
      & 0.29 & -85\% & 0.19 & -72\% & 0.40 & -72\% & 0.48 & -22\%
      & 0.60 & -63\% \\
    \rowcolor{golden!20}+ \textsc{\methodname-AR}
      & 0.71 & -81\% & 0.84 & -81\% & 0.88 & -82\% & 0.77 & -75\% & 0.72 & -10\%
      & 0.30 & -84\% & 0.15 & -82\% & 0.39 & -81\% & 0.44 & -25\%
      & 0.58 & -67\% \\
    \midrule

    InternVL3-8B
      & 0.84 &  & 0.86 &  & 0.92 &  & 0.85 &  & 0.79 &
      & 0.56 &  & 0.68 &  & 0.72 &  & 0.69 &
      & 0.77 & \\

    \rowcolor{golden!20}+ \textsc{\methodname{}}
      & 0.84 & -66\% & 0.86 & -68\% & 0.92 & -69\% & 0.85 & -70\% & 0.79 & -44\%
      & 0.56 & -86\% & 0.68 & -82\% & 0.74 & -72\% & 0.69 & -22\%
      & 0.77 & -64\% \\

    \rowcolor{golden!20}+ \textsc{\methodname-AR}
      & 0.84 & -86\% & 0.86 & -81\% & 0.92 & -80\% & 0.85 & -78\% & 0.72 & -84\%
      & 0.55 & -85\% & 0.68 & -82\% & 0.74 & -81\% & 0.68 & -31\%
      & 0.76 & -76\% \\
      
    \midrule

    Qwen2.5-VL-72B
      & 0.87 & & 0.87 & & 0.96 & & 0.75 & & 0.81 &
      & 0.62 &  & 0.77 &  & 0.73 &  & 0.74 &
      & 0.79 & \\

    \rowcolor{golden!20}+ \textsc{\methodname{}}
      & 0.87 & -85\% & 0.84 & -77\% & 0.95 & -84\% & 0.76 & -64\% & 0.79 & -77\%
      & 0.62 & -86\% & 0.79 & -82\% & 0.84 & -72\% & 0.74 & -7\%
      & 0.80 & -70\% \\
    \midrule

    GPT-4o
      & 0.78 &  & 0.56 &  & 0.80 &  & 0.77 &  & 0.76 &
      & 0.57 &  & 0.61 &  & 0.75 &  & 0.64 &
      & 0.69 &  \\

    \rowcolor{golden!20}+ \textsc{\methodname{}}
      & 0.78 & -60\% & 0.56 & -60\% & 0.80 & -36\% & 0.75 & -33\% & 0.75 & -47\%
      & 0.56 & -85\% & 0.61 & -84\% & 0.73 & -76\% & 0.61 & -17\%
      & 0.68 & -55\% \\

    \rowcolor{golden!20}+ \textsc{\methodname{}-AR}
      & 0.74 & -85\% & 0.52 & -85\% & 0.78 & -88\% & 0.73 & -84\% & 0.71 & -82\%
      & 0.56 & -85\% & 0.62 & -84\% & 0.71 & -82\% & 0.58 & -28\%
      & 0.66 & -78\% \\

    \bottomrule
  \end{tabular}}
  \caption{\textbf{Benchmark performance} and estimated prefill-stage savings for \textbf{Cost} (measured in FLOPS for local models or \$ for API models). Reporting CARES-AR for auto-regressive prediction. CARES was trained on document data, where other domain datasets shows similar performance with less aggressive Cost saved.}
  \label{tab:main}
\end{table*}

\section{Results \& Analysis}
\label{sec:exp}



This section presents the experimental evaluation of \methodname{}. We begin by describing the benchmarks and evaluation metrics (\ref{subsec::benchmarks}), followed by the main results (\ref{subsec::results}), and finally a comprehensive ablation study (\ref{subsec::ablations}).

\subsection{Experimental Setup}
\label{subsec::benchmarks}
\noindent\textbf{Training Data}
To train the resolution selector, we construct a dataset of images and queries $(x,q)$ we automatically annotated with the minimal sufficient resolution $r^{\star}$.
We construct an 80K-sample training set by randomly sampling 20K instances from each of four datasets: TextVQA \cite{TextVQA}, ChartQA \cite{chartqa}, DocVQA \cite{docvqa}, and LLaVA-Multi \cite{jiang2024mantis}, covering documents and natural images domains.

\noindent\textbf{Training details}
We train \methodname{} on the curated data described in \ref{subsec::dataset_generation} for 6 epochs using a learning rate of $1e-3$ and a batch size of 32.
We optimize the standard cross-entropy loss over the fixed resolution labels:
\[
\mathcal{L}(\theta)=\mathrm{CE}\Big(f_\theta(z),\,r^\star\Big).
\]
Where $f_\theta(z)$ is \methodname{} composed of a frozen VLM and the lightweight classifier.
In addition, we apply label smoothing of 0.05 to support continuous resolutions at inference time.

\paragraph{VLM variant training details.}
For the autoregressive (AR) Granite-Docling instantiation, we use the same training set and the same discrete supervision labels. The model is fine-tuned with LoRA of rank 8, while the base model remains frozen. Training is performed with next-token supervision over the resolution tokens, and for efficiency, generation length is set to 1. Learning rate is set to $1e-5$ and a batch size of 64 for 3 epochs.

\begin{figure}[bh]
\centering
\includegraphics[width=\linewidth,]{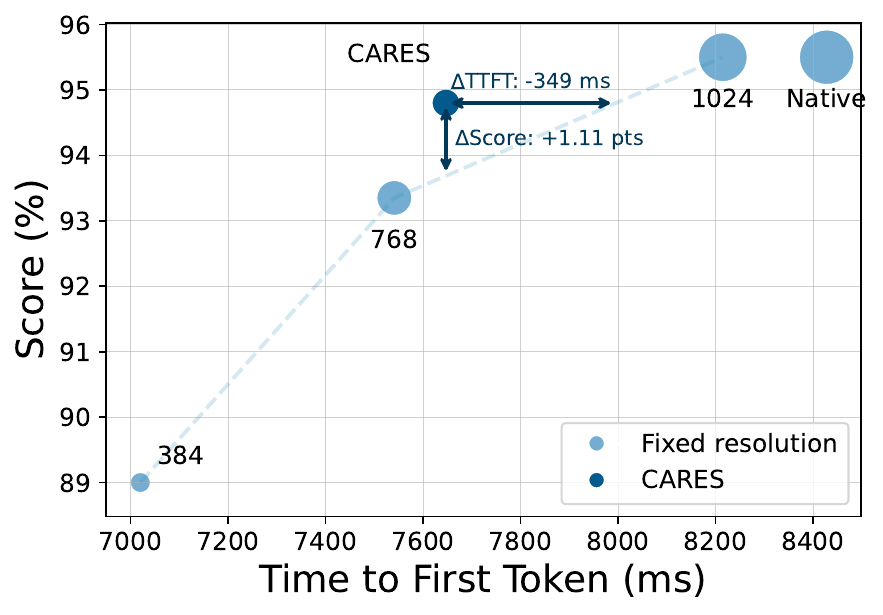}
\caption{Accuracy vs. TTFT for DocVQA with Qwen2.5-VL-72B across native and fixed-resolution settings versus CARES. Bubble size indicates the number of pixels processed by the model.} 
\label{fig:qwen_tradeoff}
\end{figure}

\begin{figure}[bh]
\centering
\begin{subfigure}[b]{0.45\textwidth}
    \centering
    \includegraphics[width=\linewidth]{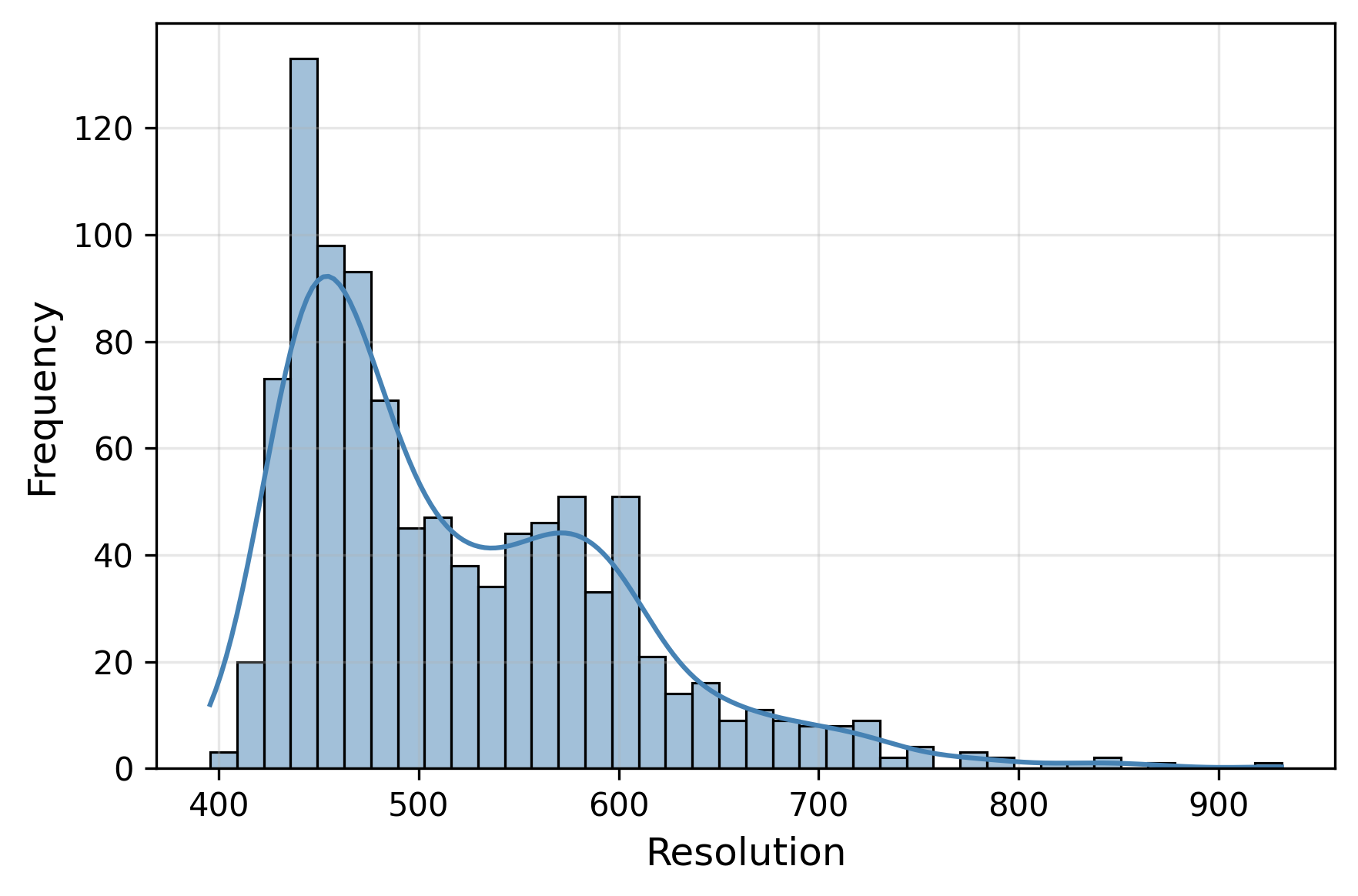}
\end{subfigure}
\caption{Histogram of the predicted resolutions $\tilde r$ by \methodname{} for OCRBench.} 
\label{fig:histogram_ocr}
\end{figure}

\noindent\textbf{Evaluation}
We evaluate on nine public benchmarks varying from documents to natural images: {Ai2D} \cite{ai2d}, {ChartQA} \cite{chartqa}, {DocVQA} \cite{docvqa}, {OCRBench} \cite{ocrbench}, and {SeedBench‑2} \cite{li2023seed2}, {MMMU} \cite{yue2023mmmu}, {RealWorldQA} \cite{xai2024realworldqa}, {InfoVQA} \cite{mathew2022infographicvqa} and {MathVista} \cite{lu2024mathvista}.
For Ai2D, ChartQA, and SeedBench-2 we report exact-match accuracy. For DocVQA and OCRBench we report Average Normalized Levenshtein Similarity (ANLS). 
All evaluations were performed with the standard lmms-eval \cite{zhang2024lmmsevalrealitycheckevaluation} setup.
We also report a macro-averaged Performance (\%) across all datasets.

\subsection{Main results}
\label{subsec::results}

We evaluate \methodname{} across \textbf{Granite-Vision 3.3-2B} \cite{team2025granite}, \textbf{InternVL3-8B} \cite{zhu2025internvl3}, \textbf{Qwen2.5-VL-72B} \cite{Qwen2.5-VL}, and \textbf{GPT-4o} \cite{achiam2023gpt}. 
We also report prefill-stage FLOPS savings for locally run models, and estimated dollar savings in API usage for GPT-4o.
As summarized in Table~\ref{tab:main}, \methodname{} maintains accuracy while cutting prefill compute: averaged over models and datasets, prefill FLOPs drop by \textbf{65--85\%} with at most a sub-point change in macro performance relative to always using the highest/native resolution. The effect is consistent for compact (Granite-Vision 3.3-2B) and large (Qwen2.5-VL-72B) backbones, and holds for GPT-4o accessed via API (accuracy parity at comparable quality). 

Fig.~\ref{fig:qwen_tradeoff} shows the accuracy–latency frontier: \methodname{} matches near-native accuracy while using far fewer TFLOPs (e.g., $2.58$ vs.\ $7.5$) and achieving $\sim 1$ second lower time-to-first-token (TTFT); static high-res inputs (e.g., $1024^2$) incur substantial compute with limited TTFT gains, whereas fixed low-res ($384^2$) improves TTFT at the cost of quality. The query-aware routing yields a superior Pareto point.

Finally, the distribution of predicted continuous resolutions $\tilde r$ (Fig.~\ref{fig:histogram_ocr}) and the comparison in Table~\ref{tab:discrete_vs_cont} indicate that continuous routing adapts per instance, matches or slightly improves accuracy over a discrete menu, and saves additional compute without quality loss.

\subsection{Cross-Teacher Agreement for Resolution Labels}
\label{app:teacher-agreement}

Because our supervision is generated by rolling out a pretrained VLM at multiple resolutions, one natural question is whether the resulting labels depend strongly on the specific annotating model. To test this, we compare labels generated by two substantially different teachers: \textbf{Granite-Vision-2B} and \textbf{Qwen3-VL-235B}, on a shared subset of 1000 examples.

\begin{figure}
    \centering
    \includegraphics[width=\linewidth]{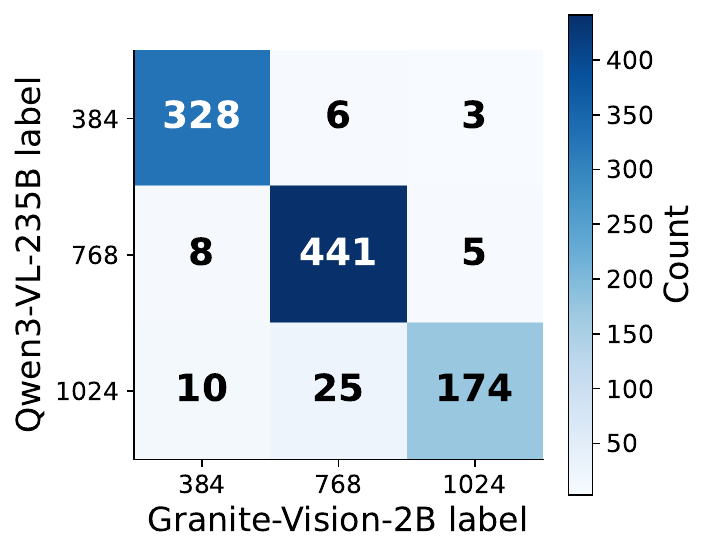}
    \caption{Confusion matrix between sufficient-resolution labels generated by Granite-Vision-2B and Qwen3-VL-235B on a shared subset. Most mass lies on the diagonal, indicating strong agreement across teachers.}
    \label{fig:teacher-confusion}
\end{figure}

We find a high degree of agreement between the two annotators. The two teachers predict the same sufficient resolution for more than \textbf{95\%} of examples, with Pearson correlation \textbf{0.908} and mutual information \textbf{1.116} between their predicted sufficiency levels. The confusion matrix is shown in Table~\ref{fig:teacher-confusion}. These results suggest that the notion of sufficient resolution is largely shared across architectures and scales, and is not tied to a single model family.

This result complements the downstream transfer results in the main paper, where a selector trained using labels derived from one setup transfers well across multiple target VLMs. Together, these findings support the view that CARES captures a broadly shared notion of \emph{resolution adequacy}, rather than overfitting to one teacher's idiosyncrasies.

\subsection{Ablation study}
\label{subsec::ablations}
We conduct a series of ablations to isolate the effect of key training design choices on resolution selection accuracy and downstream benchmark performance.

\paragraph{Feature extractor.}
We ablate several frozen backbones used for feature extraction in \methodname{}, varying both model type and layer depth. As shown in Table~\ref{tab:feature_extractor}, both Qwen2.5-3B and SmolVLM achieve higher accuracy when using intermediate-layer features, outperforming their own final-layer variants. This aligns with prior findings suggesting that intermediate representations in VLMs often encode richer signals than final outputs.

Qwen2.5-3B and SmolVLM both process the image and query jointly within a unified transformer, in contrast to SigLIP v2’s dual-encoder architecture, where vision and language are encoded separately. For SigLIP, we follow the original design by pooling the outputs of each tower, concatenating them, and passing the result to the classifier head. While this setup is architecturally simple, it underperforms joint encoding by a considerable margin (56.1\% accuracy), and it requires more parameters than the lightweight SmolVLM.

Although Qwen2.5-3B achieves the best overall accuracy, we adopt SmolVLM as our default backbone due to its favorable trade-off between performance, size, and efficiency, making it a more practical choice for real-world pre-processing.

\begin{table}[ht]
    \centering
    \begin{tabular}{lllc}
    \toprule
        Model & Layer & Params & Accuracy \\
        \midrule
        SigLIP v2 & - & 0.8B & 56.1\% \\
        SmolVLM & Mid & 0.35B & 63.3\% \\
        SmolVLM & Last & 0.5B & 62.3\% \\
        Qwen2.5\textendash3B & Mid & 2.3B & 67.2\% \\
        Qwen2.5\textendash3B & Last & 3.75B & 66.2\% \\
    \bottomrule
    \end{tabular}
    \caption{\textbf{Feature extractor.}
        Validation accuracy and parameter count for different frozen feature extractors used in \methodname{}. All models are trained to classify among three resolution choices. For SmolVLM and Qwen2.5-3B, we compare features extracted from intermediate (\textsc{Mid}) and final (\textsc{Last}) layers. For SigLIP, the pooled outputs from the vision and language towers are concatenated and passed to the classifier head. Qwen2.5-3B provides the best performance, while SmolVLM offers strong accuracy with minimal size.
    }
    \label{tab:feature_extractor}
\end{table}

\paragraph{Resolution menu size.}
We compare training with binary $\mathcal{R}_d=\{384,1024\}$ ($|\mathcal{R}_d|=2$) vs. ternary $\mathcal{R}_d=\{384,768,1024\}$ ($|\mathcal{R}_d|=3$) resolution choices. Table \ref{tab:binary_vs_ternary} reports both the classification accuracy and the downstream performance of Granite Vision, averaged over 5 benchmarks. As expected, the two-way classification yields higher validation accuracy in the resolution classification task compared to the more challenging three-way classification. But the ternary setup leads to better downstream benchmark performance due to the finer-grained control.

\begin{table}[ht]
    \centering
    \begin{tabular}{ccc}
    \toprule
         & Resolution  & Downstream \\
        $|\mathcal{R}_d|$ &  Accuracy &  Accuracy\\
        \midrule
        2 & 96.2\% & 0.76 \\
        3 & 67.2\% & 0.80 \\
        \bottomrule
    \end{tabular}
    \caption{
\textbf{Binary vs. Ternary Resolution Classification.}
We compare binary ($|\mathcal{R}_d|=2$, using $\{384,1024\}$) and ternary ($|\mathcal{R}_d|=3$, using $\{384,768,1024\}$) resolution selection setups. The binary classifier achieves higher accuracy on the resolution prediction task due to its reduced complexity, while the ternary classifier improves downstream performance by enabling finer control over resolution. Reported downstream accuracy is averaged over 5 vision-language benchmarks using Granite Vision.
}
    \label{tab:binary_vs_ternary}
\end{table}

\paragraph{Discrete vs. continuous.}
\methodname{} is trained as a discrete resolution classifier, but at inference time, it can produce either discrete predictions or a continuous estimate via interpolation. In Table~\ref{tab:discrete_vs_cont}, we compare the impact of discrete versus continuous inference across three VLM backbones. All scores and FLOPS deltas are averaged over nine benchmarks. We find that continuous resolution selection achieves comparable accuracy to both discrete and native strategies, while significantly reducing compute. For example, with Granite-Vision 3.3-2B and InternVL3-8B, FLOPS are reduced by 63\% using continuous prediction, compared to 46\% with discrete. These results suggest that continuous inference allows finer control over input resolution and leads to more efficient inference without compromising performance.

\begin{table}[ht]
  \centering
  \small

  \begin{tabular}{l l c r}
    \toprule
    \textbf{Model} & \textbf{Resolution} & \textbf{Score} & \textbf{FLOPS} \\
    \midrule
    Granite-Vision 3.3-2B & Native      & 0.803  &    \\
                          & Discrete    & 0.801 & -46\% \\
                          & Continuous  & 0.804  & -63\% \\
    \midrule
    InternVL3-8B          & Native      & 0.851 &    \\
                          & Discrete    & 0.851 & -46\% \\
                          & Continuous  & 0.851 & -63\% \\
    \midrule
    Qwen2.5-VL-72B        & Native      & 0.851 &    \\
                          & Discrete    & 0.852 & -74\% \\
                          & Continuous  & 0.839 & -80\% \\
    \bottomrule
  \end{tabular}
    \caption{\textbf{Discrete vs. Continuous Resolution Selector.} . The overall score and relative FLOPS delta per resolution strategy are averaged over 5 benchmarks. Using continuous resolutions allows finer control of the resolution, resulting in a lower resolution and computation with no drop in accuracy.}
  \label{tab:discrete_vs_cont}
\end{table}

\paragraph{Label smoothing.}

To bridge the mismatch between \emph{discrete} supervision and our \emph{continuous} inference policy, we apply label smoothing when training the classifier over $\mathcal{R}_d$. Smoothing softens class boundaries and discourages over-confident logits, yielding better-calibrated probability distributions $p$ that are subsequently mapped to a scalar resolution via expectation (Eq.~\ref{eq:linear_expectation}).
This improves the stability of the continuous selector, reduces spurious hard escalations near decision thresholds, and translates to higher downstream utility at similar—or lower—compute. Empirically, Table~\ref{tab:smooth} shows that adding label smoothing improves OCRBench performance for Qwen2.5-VL-7B (0.821 vs.\ 0.811) while slightly \emph{reducing} expected FLOPS, supporting its role as a simple but effective regularizer for continuous-resolution deployment.

\begin{table}[bh]
  \centering
  \small
  \resizebox{0.45\textwidth}{!}{\begin{tabular}{ l c r}
    \toprule
     \textbf{Setting} & \textbf{Score} & \textbf{FLOPS} \\
    \midrule
      Native resolution     & 0.824 &      \\
        CARES Without label-smoothing & 0.811 & -60.5\% \\
        CARES With label-smoothing    & 0.821 & -63.8\% \\
    \bottomrule
  \end{tabular}}
  \caption{\textbf{Label smoothing effect.} Evaluated on OCRBench with Qwen2.5-VL-7B. Comparison of native resolution and training with or without label smoothing. FLOPs indicate relative change.}
  \label{tab:smooth}
\end{table}

\section{Discussion and Conclusion}

Inference efficiency has become a critical concern for modern vision-language systems. Most user queries do not require high-resolution inputs, yet current deployments often process all images at native or tiled resolutions by default. This leads to bloated token counts, slower response times, and higher costs. CARES addresses this challenge with a lightweight, model-agnostic approach that dynamically selects input resolution based on the query. By acting before tokenization, it provides a clean and practical lever for controlling inference cost while maintaining output quality.

\paragraph{Key Takeaways}
\begin{itemize}[leftmargin=*, itemsep=1pt, topsep=1pt, parsep=1pt]
    \item CARES reduces compute and latency across a wide range of models and benchmarks, with minimal to no loss in task accuracy.
    
    \item It requires no changes to the vision-language model and works as a plug-in component, making it easy to integrate into real-world pipelines.
    
    \item CARES adapts resolution based on the specific query, using a single low-cost pass to determine how much visual detail is needed.
        
    \item The design is compact and efficient, enabling wide applicability without adding large overhead to the main model.
\end{itemize}



\noindent
Overall, CARES highlights the value of adaptive pixel allocation as a simple yet powerful strategy for efficient multimodal inference. It complements existing techniques for token-level compression and opens up a new path for practical deployment of vision-language models at scale.

\section*{Limitations}
CARES depends on a frozen proxy VLM for low-resolution features; domains requiring extremely fine cues (e.g., dense OCR, medical imagery) may be under-allocated. Our supervision uses multi-resolution rollouts of a target VLM and thus inherits that model’s biases and limited language support. Robustness to model perturbation at inference \cite{galil2026maximal} or noise in annotations \cite{kimhi2025noisyannotationssemanticsegmentation} are not explored. We evaluate single-image, single-turn inputs only; multi-image, video, streaming, and joint resolution–tiling selection are left to future work. We do not study safety, robustness to adversarial prompts, or detailed cost–latency trade-offs across hardware.

\clearpage

\bibliography{custom}
\newpage

\appendix
\newpage
\newpage
\section{Additional Analysis and Results}
\label{app:additional_results}

This appendix provides additional qualitative and quantitative analysis of \methodname{}.

\subsection{Extended token count evaluation}
\label{app:token_count}

\subsubsection{Textual token statistics}
\label{app:text_tokens}

We analyze the number of textual tokens across the evaluated benchmarks using the Qwen2.5-VL tokenizer. For each dataset, we compute the average number of tokens in the full input prompt, including the question, answer choices (when applicable), and the instruction suffix (e.g., \emph{``Answer with the option's letter...''}). The results are summarized in Table~\ref{tab:avg_tokens}.

We observe that most benchmarks contain relatively short textual inputs, typically in the range of 20--50 tokens. The longest prompts appear in MMMU, with an average of approximately 100 tokens, due to its multi-choice and instruction-heavy format.

Based on this analysis, we adopt a fixed text length of $T=100$ tokens in our token composition study (Section~\ref{app:token_count}). This choice reflects a conservative upper bound over the evaluated benchmarks and ensures that our analysis does not underestimate the contribution of textual tokens. In practice, this assumption is favorable to text, as most datasets contain substantially fewer tokens, further increasing the relative dominance of visual tokens in real-world settings.

For a fixed text prompt of $T=100$ tokens, the visual-token fraction is
\[
P(V;T)=100\cdot \frac{V}{V+T}.
\]

For Qwen2.5-VL, we approximate the number of LLM-side visual tokens by
\[
V_{\text{Qwen}}(H,W)=\left\lceil \frac{H}{28}\right\rceil \left\lceil \frac{W}{28}\right\rceil,
\]
reflecting patch size 14 with spatial merge size 2.

For AnyRes tiled models (e.g., LLaVA-NeXT / Granite-Vision \cite{granite-4.0-3b-vision} processing), a square-input abstraction is
\[
V_{\text{AnyRes}}(s)=576\Bigl(1+\min\{\lceil s/336\rceil^2,4\}\Bigr),
\]
where 576 is the per-image embedding length and the default square grid saturates at a $2\times2$ local tiling plus a global view.

For InternVL3, we use a tile-based approximation consistent with its dynamic-resolution preprocessing:
$$
V_{\text{Intern}}(s)=256\Bigl(n(s)+\mathbf{1}[n(s)>1]\Bigr),$$
$$n(s)=\min\!\left(\left\lceil \frac{s}{448}\right\rceil^2,40\right),$$
where each 448$\times$448 tile contributes 256 LLM-side tokens after pixel unshuffle, and an additional thumbnail is used when more than one tile is present.

These expressions explain the different regimes in Table~\ref{tab:tokens_full}: Qwen2.5-VL grows approximately quadratically with resolution, AnyRes saturates early due to a fixed tile budget, and InternVL3 follows a stepwise tile-based growth pattern.

\begin{table}[t]
\centering
\small
\resizebox{\linewidth}{!}{\begin{tabular}{c|l|c|c|c}
\hline
\textbf{Resolution} & \textbf{Model} & \textbf{Visual} & \textbf{Total} & \textbf{\% Visual} \\ \hline
\multirow{3}{*}{336$\times$336}
& AnyRes (Tiled) & 1152 & 1252 & 92.0\% \\
& Qwen2.5-VL     & 144  & 244  & 59.0\% \\
& InternVL3      & 256  & 356  & 71.9\% \\ \hline

\multirow{3}{*}{672$\times$672}
& AnyRes (Tiled) & 2880 & 2980 & 96.6\% \\
& Qwen2.5-VL     & 576  & 676  & 85.2\% \\
& InternVL3      & 1280 & 1380 & 92.8\% \\ \hline

\multirow{3}{*}{1024$\times$1024}
& AnyRes (Tiled) & 2880 & 2980 & 96.6\% \\
& Qwen2.5-VL     & 1369 & 1469 & 93.2\% \\
& InternVL3      & 2560 & 2660 & 96.2\% \\ \hline

\multirow{3}{*}{2048$\times$2048}
& AnyRes (Tiled) & 2880 & 2980 & 96.6\% \\
& Qwen2.5-VL     & 5476 & 5576 & 98.2\% \\
& InternVL3      & 6656 & 6756 & 98.5\% \\ \hline

\multirow{3}{*}{4096$\times$4096}
& AnyRes (Tiled) & 2880  & 2980  & 96.6\% \\
& Qwen2.5-VL     & 21609 & 21709 & 99.5\% \\
& InternVL3      & 10496 & 10596 & 99.1\% \\ \hline
\end{tabular}}
\caption{Visual vs.\ text token composition (assuming 100 text tokens) across representative VLM architectures. Qwen2.5-VL scales approximately quadratically with input size, AnyRes-style tiled models saturate due to a fixed square tiling budget, and InternVL3 follows a dynamic tile-based scaling with 448$\times$448 tiles and an additional thumbnail view.}
\label{tab:tokens_full}
\end{table}

\begin{table}[t]
\centering
\small
\setlength{\tabcolsep}{6pt}
\begin{tabular}{lc}
\toprule
\textbf{Dataset} & \textbf{Avg. Tokens} \\
\midrule
AI2D        & 44.3 \\
ChartQA     & 21.6 \\
DocVQA      & 21.2 \\
OCRBench    & 22.5 \\
SeedBench-2 & 45.8 \\
MMMU        & 100.4 \\
RealWorldQA & 37.5 \\
InfoVQA     & 24.0 \\
MathVista   & 67.3 \\
\bottomrule
\end{tabular}
\caption{\textbf{Average number of textual tokens per benchmark.} Token counts are computed using the Qwen2.5-VL tokenizer and include the full input prompt (question, answer choices when applicable, and instruction suffix such as \emph{``Answer with the option's letter...''}).}
\label{tab:avg_tokens}
\end{table}

\subsection{Label generation pipeline}
\label{app:label_pipeline}

\begin{figure}[b]
\centering
\includegraphics[width=0.8\linewidth,]{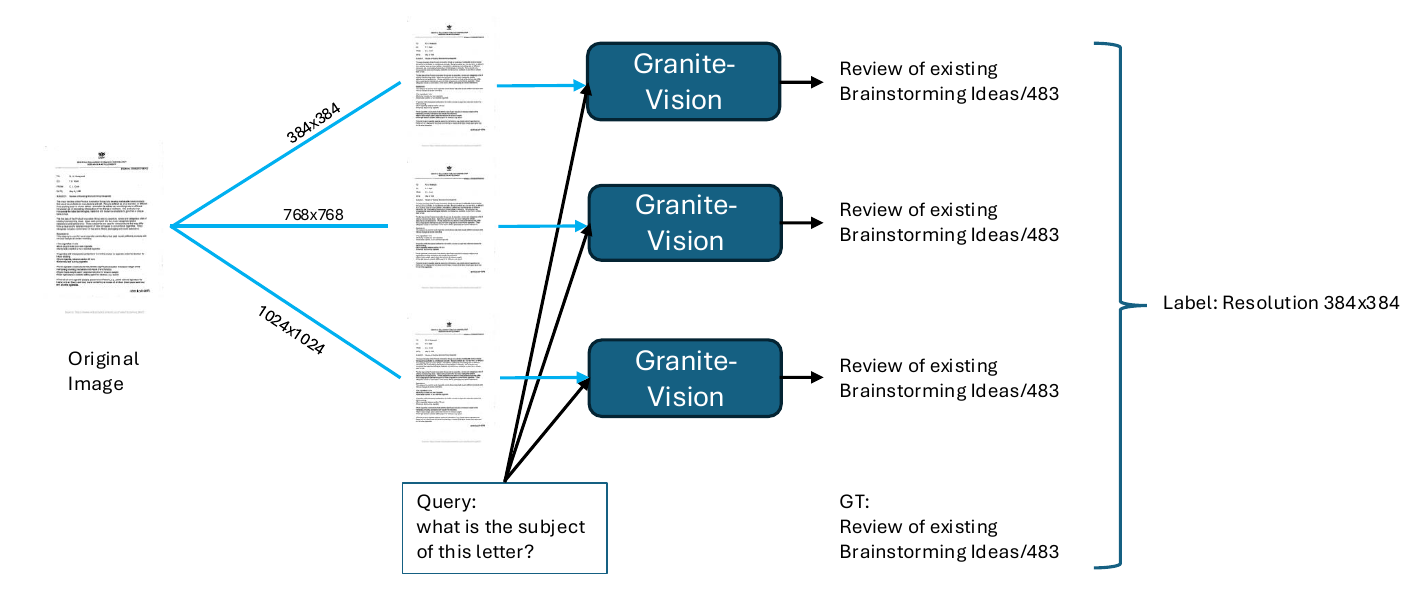}
\caption{\textbf{Label generation pipeline for training \methodname{}.} For each image-query pair, we evaluate a pretrained VLM at multiple fixed resolutions and assign the smallest resolution that satisfies the sufficiency criterion as the supervision label.}\label{fig:data_pipeline}
\end{figure}

Figure~\ref{fig:data_pipeline} illustrates the supervision pipeline used to train \methodname{}. For each image-query pair, we evaluate a pretrained VLM at several fixed resolutions and compare its prediction to the ground-truth answer. The smallest resolution whose score satisfies the sufficiency criterion is used as the training target. This process transforms downstream task behavior into per-example supervision for resolution selection, enabling \methodname{} to learn when higher visual detail is genuinely needed.

\subsection{Adaptive Selection vs.\ Fixed-Resolution Baselines}
\label{app:fixed-baselines}

To disentangle the benefit of adaptive resolution selection from the general robustness of VLMs to downscaling, we compare CARES against fixed-resolution baselines. Table~\ref{tab:fixed-resolution-baselines} reports an example comparison using [MODEL / benchmark setting], where all inputs are processed at a single fixed resolution.

\begin{table}[t]
\centering
\small
\begin{tabular}{l c}
\toprule
\textbf{Setting} & \textbf{Accuracy} \\
\midrule
Native          & 95.50 \\
1024            & 95.50 \\
768             & 93.35 \\
384             & 89.00 \\
CARES           & 94.80 \\
\bottomrule
\end{tabular}
\caption{Comparison between fixed-resolution inference and CARES. While naive downscaling reduces accuracy, CARES recovers most of the native performance at much lower average compute.}
\label{tab:fixed-resolution-baselines}
\end{table}

These results clarify that the gains are not simply due to the model tolerating smaller images. Rather, the adaptive policy selectively preserves high resolution for the examples that need it, while routing easier cases to much smaller inputs. This explains why CARES achieves a better accuracy--efficiency trade-off than any single fixed-resolution operating point.

\subsection{Time-to-first-token analysis}
\label{app:ttft}

Table~\ref{tab:ttft} reports time-to-first-token (TTFT) on DocVQA for representative downstream VLMs. The results mirror the FLOPS trends in the main paper: lower resolutions substantially reduce latency, while \methodname{} achieves a favorable trade-off by approaching the latency of low-resolution inference without incurring the accuracy loss of always using a small input. In particular, CARES significantly improves TTFT relative to native or fixed high-resolution processing, confirming that adaptive resolution selection translates into practical end-to-end inference gains.
\begin{table}
  \centering
  \small
  \setlength{\tabcolsep}{5pt}
\caption{\textbf{Time to First Token (TTFT, ms)} measured on H100 with batch size $1$, averaged over 100 DocVQA examples. \emph{Native} denotes the model's default input pipeline. CARES reduces TTFT substantially compared to native and fixed high-resolution settings while preserving strong downstream accuracy.}  \label{tab:ttft}
  \resizebox{\linewidth}{!}{\begin{tabular}{
    l
    S[table-format=3.1]
    S[table-format=3.1]
    S[table-format=3.1]
    S[table-format=3.1]
    S[table-format=3.1]
  }
    \toprule
    \textbf{Model} & {\textbf{Native}} & {\boldmath$1024^2$} & {\boldmath$768^2$} & {\boldmath$384^2$} & {\textbf{\methodname{}}} \\
    \midrule
    Qwen2.5-VL-7B & 435.7 & 433.8 & 220 &  76.12 & 270.1 \\
    Granite-Vision 3.3-2B & 228.6 & 201.3 & 140.1 &  96.1 & 108.9 \\
    \bottomrule
  \end{tabular}}
\end{table}

\subsection{Robustness to Proxy--Target Feature Mismatch}
\label{app:proxy-target-mismatch} 
Since CARES uses a proxy VLM to extract low-resolution image-query features, one may worry that mismatch between the proxy representation and the downstream target model could introduce bias. To directly test this, we evaluate CARES on \textbf{Qwen2.5-VL-3B} using two feature extractors: (i) features from the same target-family model, and (ii) features from \textbf{SmolVLM}, a substantially smaller proxy.

Table~\ref{tab:proxy-mismatch} shows that the two variants perform similarly across all tested benchmarks, with differences that are small relative to the benchmark scale. This indicates that the resolution-selection decision depends primarily on coarse visual-textual cues that are preserved across different VLMs, rather than requiring tight alignment between proxy and target feature spaces.

\begin{table}[t]
\centering
\small
\begin{tabular}{lcc}
\toprule
\textbf{Task} & \textbf{Qwen features} & \textbf{SmolVLM features} \\
\midrule
Ai2D      & 0.7830 $\pm$ 0.0074 & 0.7824 $\pm$ 0.0074 \\
ChartQA   & 0.8164 $\pm$ 0.0077 & 0.8080 $\pm$ 0.0079 \\
DocVQA    & 0.8814 $\pm$ 0.0044 & 0.8640 $\pm$ 0.0040 \\
OCRBench  & 0.7500 $\pm$ 0.0001 & 0.7600 $\pm$ 0.0003 \\
\bottomrule
\end{tabular}
\caption{Downstream performance of CARES when using same-family versus proxy-family features. The small differences suggest that CARES is robust to moderate proxy--target representation mismatch.}
\label{tab:proxy-mismatch}
\end{table}

\subsection{Predicted resolution distributions}
\label{app:histograms}

\begin{figure}
\centering
\begin{subfigure}[b]{0.45\textwidth}
    \centering
    \includegraphics[width=\linewidth]{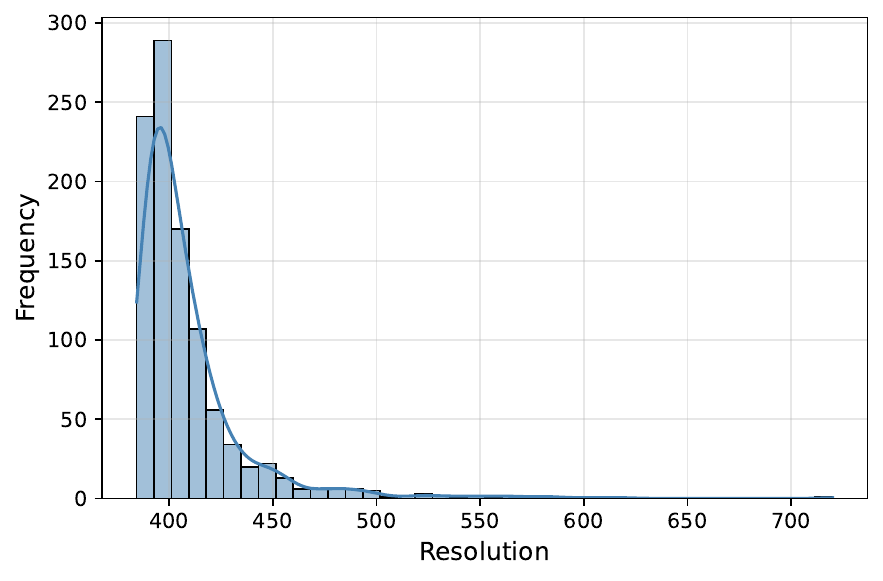}
    \caption{Ai2D.}
    \label{fig:hist_ai2d}
\end{subfigure}
\begin{subfigure}[b]{0.45\textwidth}
    \centering
    \includegraphics[width=\linewidth]{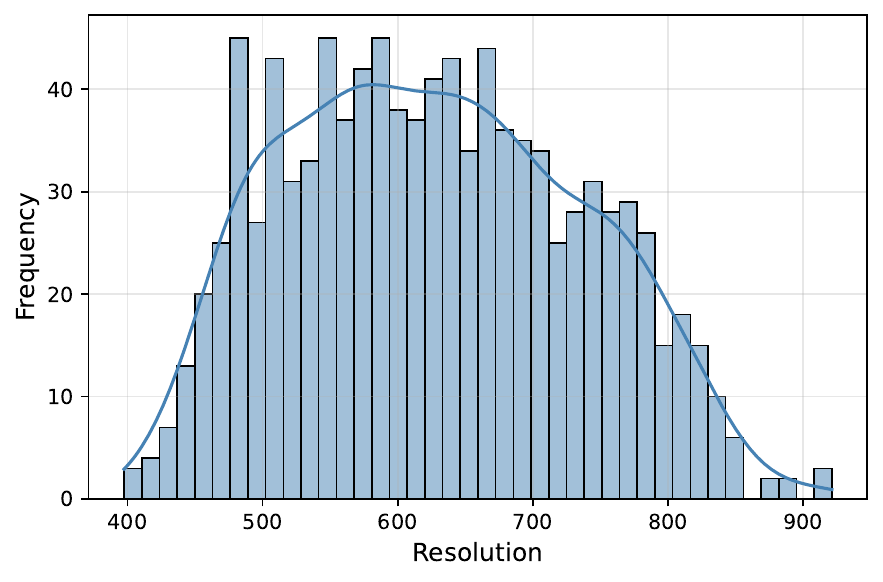}
    \caption{DocVQA.}
    \label{fig:hist_docvqa}
    \end{subfigure}
\begin{subfigure}[b]{0.45\textwidth}
    \centering
    \includegraphics[width=\linewidth]{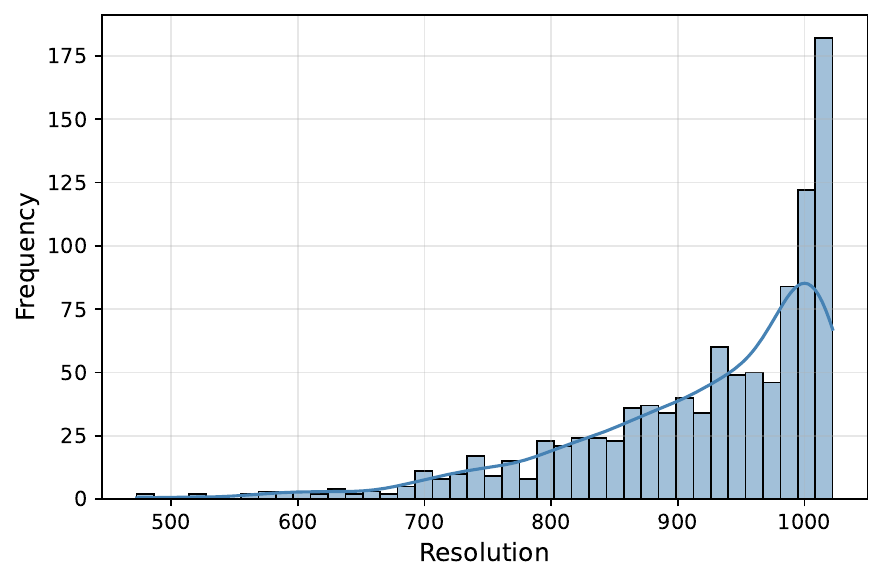}
    \caption{SeedBench-2.}
    \label{fig:hist_seed}
\end{subfigure}
\begin{subfigure}[b]{0.45\textwidth}
    \centering
    \includegraphics[width=\linewidth]{figs/ocr_hist.png}
    \caption{OCRBench.}
    \label{fig:hist_ocr}
\end{subfigure}
\caption{\textbf{Histograms of the predicted continuous resolutions $\tilde r$ by \methodname{}.} CARES routes many Ai2D examples to lower resolutions, while SeedBench-2 shifts toward higher resolutions. DocVQA and OCRBench show broader distributions, reflecting their mixture of coarse and fine-grained queries, including dense text and complex layouts.}\label{fig:histograms_all}
\end{figure}

Figure~\ref{fig:histograms_all} shows the distribution of continuous resolutions predicted by \methodname{} across different benchmarks. The histograms highlight that the selector adapts its behavior to the underlying task: Ai2D is dominated by lower-resolution predictions, suggesting that many diagram-understanding questions require only coarse visual information; SeedBench-2 shifts toward higher resolutions, reflecting the need for finer-grained visual recognition; and DocVQA and OCRBench exhibit broader distributions, indicating a mixture of easy and detail-sensitive examples. This behavior is consistent with the intended design of \methodname{}, which escalates resolution only when the image-query pair appears to demand additional visual detail.

\end{document}